\documentclass{article}
\usepackage{ijcai20}

\usepackage{times}
\usepackage{soul}
\usepackage{url}
\usepackage[hidelinks]{hyperref}
\usepackage[utf8]{inputenc}
\usepackage[small]{caption}
\usepackage{graphicx}
\usepackage{amsmath}
\usepackage{amsthm}
\usepackage{booktabs}
\usepackage{algorithm}
\usepackage{algorithmic}
\usepackage{verbatim}
\usepackage{tablefootnote}
\usepackage{titling}

\newcommand{\A}{\mathcal{A}}
\newcommand{\Sp}{\mathcal{S}}

\title{CoordiQ : Coordinated Q-learning for Electric Vehicle Charging Recommendation}

\author{
    Carter W. Blum$^{1,2}$\thanks{Contact Author. blumx116@umn.edu}\quad\quad
    Liu Hao$^1$\quad\quad
    Hui Xiong$^{1,3}$\\\\
    $^1$ Business Intelligence Lab, Baidu Inc., Beijing, CN\\
    $^2$ University of Minnesota, Minneapolis, MN, US\\
    $^3$ Rutgers University, Newark, NJ, US
    }
\date{December 2019}

\begin{document}

\maketitle

\section{Abstract}

Electric vehicles have been rapidly increasing in usage, but stations to charge them have not always kept up with demand, so efficient routing of vehicles to stations is critical to operating at maximum efficiency.
Deciding which stations to recommend drivers to is a complex problem with a multitude of possible recommendations, volatile usage patterns and temporally extended consequences of recommendations.
Reinforcement learning offers a powerful paradigm for solving sequential decision-making problems, but traditional methods may struggle with sample efficiency due to the high number of possible actions.
By developing a model that allows complex representations of actions, we improve outcomes for users of our system by over 30\% when compared to existing baselines in a simulation.
If implemented widely, these better recommendations can globally save over 4 million person-hours of waiting and driving each year.

\section{Introduction}
As the public becomes more ecologically conscious, electric vehicles are surging in popularity, putting up double digit growth rates year over year.
However, despite consumer enthusiasm, electric vehicles still face some major hurdles that they must face before they can truly become mainstream.
One such hurdle is public charging stations. 
Because not all electric vehicle owners have the infrastructure to charge their vehicles at home, many rely on electric vehicle charging stations to keep their tank full.
This problem is particularly prevalent among electric car owners in urban environments, who often do not have a space to charge their own vehicle. 

Unfortunately, electric vehicle charging stations have not spread as prolifically as the vehicles themselves, causing a potential lack of supply.
Each charging station has a number of chargers, each of which can service one car at a time.
Our data analysis shows that, in the city of Beijing, at least 20\% of charging stations are at 100\% occupancy - even during periods of low-usage.
Electric vehicles can take hours to charge, so stations that are full can stay full for long periods of time, resulting in very long wait times for any vehicles waiting for a charger.

In our increasingly digital world, consumers increasingly rely on mobile apps to direct them to the best utilities. Perhaps the most prominent such utility in China is Baidu Maps, which services 300 million users per month and processes hundreds of thousands of electric vehicle charging station queries per year. Consumers query the app, send it their location, and it sends them back recommendations for nearby charging stations. Current applications simply return a sorted list of the nearest highly rated stations, but this can cause problems, as these are often full, and wait time can cause more of an inconvenience to the user than driving time.

This project implements and evaluates a reinforcement learning agent to intelligently respond to queries, recommending users to nearby charging stations with minimal inconvenience. The project assumes no control over when the app  will receive queries and it assumes that there are many other electric vehicles that it has no control over. Despite these constraints, we show that the agent is able to learn effective recommendations that reduce the user's time-to-charging by over 45\%.

The key contributions of this work are as follows : 
\begin{itemize}
    \item We present the first machine learning based algorithm for charging station recommendation, and implement a context-aware reinforcement algorithm that coordinates between different recommendations. It substantially outperforms all existing baselines.
    \item To the best of our knowledge, there are no published works yet using Graph Neural Networks (GNNs) to solve reinforcement learning tasks. We propose an architecture and show it improves performance
\end{itemize}
\section{Related Works}
Reinforcement learning is a useful paradigm for solving sequential control tasks of many times. 
Reinforcement learning models operate in a Markov Decision Process, learning a probability distribution over a finite set of actions to maximize reward \cite{RusselNorvig}\cite{SuttonBarto}.
With the recent increase of processing power and the modeling power of neural networks, Deep Q-Networks (DQNs) have shown impressive capabilities to handle a diverse range of tasks \cite{Mnih1}.
To that extent, Deep Q-Networks have been shown to learn complex tasks, and have even been able to excede human level control when solving many problems \cite{MnihHuman}.

Reinforcement Learning has been successfully applied to improving charging of electric vehicles. 
However, most of the research has focused on improving the charging experience once a vehicle has arrived in a station, such as in \cite{Charging1}, \cite{Charging2} and \cite{Charging3}.
These methods successfully improve load distribution equity in stations, and schedule charging demands accordingly, as in \cite{Charging4}.
\cite{StationPlanning} also analyze charging station demand to improve the layout of stations to better handle peak demand. 

However, reinforcement learning has been applied to electric vehicle station recommendation as well. \cite{EBus} provides a thorough analysis of bus charging recommendations. 
However, the methods provided focus specifically on buses, following strict and predictable schedules, and the primary difficulty is in choosing when to charge, not where. 
The question being approached in this paper is the opposite, we assume that the user needs to charge now and the recommender attempts to minimize the inconvenience to the user.
On the other hand, \cite{RuleBasedInvalid} focuses on choosing what stations to recommend taxis to. The authors employ a rule-based method to do recommendation, but the method provided relies on future knowledge of the occupancy of a charging station at the time that the user arrives. 
Furthermore, it assigns each taxi in isolation, without accounting for the impact that one taxi's recommendation may have on the wait-time of another.

Outside of electric vehicle recommendation, there is a large body of work in spatio-temporal reinforcement learning.
In \cite{DidiBig}, the authors use a tabular method to recommend idle taxis to nearby regions and allocate dispatches. 
While this approach was extremely successful for its authors, it heavily exploits the interchangeability of taxis, a property that electric vehicle charging stations do not have to the same degree.

A non-tabular approach has also been pursue in several works, including \cite{GoodImages}, which implements a capsule network to recommend idle taxi cruising.
This approach is also modified and expanded in several similar papers, including \cite{Taxi1}, \cite{Taxi2}, \cite{Taxi3}, \cite{Taxi4}, \cite{Taxi5}.
These papers improve taxi idle cruising to reposition them to nearby areas, either by considering the taxis in concert or in isolation.
They train reinforcement learning models in simulations, with space discretized in to tiles, and the use a convolutional neural network architecture to output Q-values for each location, and then selecting the highest Q-value from the adjacent tiles.
However, most tiles are not valid actions for station recommendation, giving this method a very high noise to signal ratio when applied to charging station recommendation.

A couple other approaches to spatio-temporal recommendation have also been used.
One approach is to try to minimize the divergence between the distribution of queries and taxi locations, as demonstrated in \cite{SpatialDistribution}.
While this approach appears very promising, the number of 0 support regions in both query and station distributions causes many distributional distance metrics to be unstable.
Another direction is explored in \cite{RegionSplit}, which uses a tabular method but dynamically splits the zones whenever it can improve Q-value accuracy.
This method can be suitable for macro-planning, but charging station recommendation requires distinguishing between stations that are close together. 
Under this paradigm, those stations may be grouped in the same zone, resulting in decreased performance.

An alternate method for handling spatially related data is to use graphical neural networks.
Graphical neural networks define the input as a graph, with vertices edges and weights. 
This allows information from nodes to propagate from one to another \cite{OGGNN}.
Initial methods put restrictions on the weights of the graph convolution kernels to ensure that repeated convolution would converge \cite{GNNRecur}\cite{ChebNets}.
However, more recently methods methods have been doing away with these restrictions \cite{GCN}.
One such method that has had significant success is GraphSage, which utilizes local graph convolution over a finite set of iterations.
This method does not guaranteed converge to a set value, but can, in practice achieve similar, if not superior results with significantly less computation.
\section{Methodology}
We formulate the problem as a Markov Decision Process (MDP), with a statespace $\Sp$, action space $\A$, transition function $T$, and reward function $R$.
Our recommendation system is represented by a policy $\pi:\Sp \rightarrow \A$
At each timestep in an MDP, an agent must look at it's current state and choose an action to maximize its expected discounted rewards. 
The expected rewards of action $a \in \A$ while at state $s \in \Sp$ is given by :
\begin{equation*}
    Q(s,a) = E[R(t)] + \gamma \sum_{s' \in \Sp} T(s'|s,a) \max_{a' \in \A} Q(s',a')
\end{equation*}
here $\gamma \in [0, 1)$ is the discount factor. 

\textit{States:} In this problem, the state is the information available to the station to make recommendations with. 
Some information is available for each city that may be relevant to all stations, such as estimates of traffic, the weather, day of week, and time of day.
For a given state $s$, we'll refer to this global information as $s_G$.
For each of the $N$ stations, information about the status of its chargers is available, as well as its location and id.
The information available about the $n^{th}$ station is denoted as $s_n$.
Finally, the agent has access to information about the queries it is responding to, including their location and user id, as well as time of query.
For the $i^{th}$ query received at time $t$, the relevant state information is denoted as $s_i$.

We discretize the each region in to a grid, with each grid cell having an area of roughly $0.25$km$^2$. 
We similarly turn the system in to a sequential decision process by dividing time in to non-overlapping intervals of duration $\tau$.

\textit{Actions:} At each timestep $t$, the agent receives a set of user queries $Q_t$ for which it must provide recommendation.
The station can recommend each of the $|Q_t|$ user queries to any of the $N$ stations, resulting in a possible action space of $N^{|Q_t|}$ combinations.
This poses a problem, because not only is the possible action space potentially huge, but the available actions vary from timestep to timestep.
To address this, the agent makes each of its decisions for each query sequentially and executes all actions after decisions about all queries have been made.
When there is no loss of clarity, we will instead refer to recommending a user to station $n$ as action $a_n$.

\textit{Rewards:} After a driver is recommended to a station, they must drive there.
Subsequently, if the station is full, then they must spend time waiting for a space to open up.
To accommodate these requirements, the agent must then satisfy the multi-modal objective of minimizing both of these.
Let $t^i_{drive}$ be the driving time for the user corresponding to query $i$, as estimated by driving speed and distance. 
Similarly, we define $t^i_{wait}$ to be the the number of timesteps that the user spends waiting at the station before a spot opens for them.
As the agent wants to find an optimal split between users,
the goal is to then minimize the following equation
\begin{equation*}
     \sum_t \sum_{i \in \mathcal{Q}(t)} t^i_{wait} + \lambda t^i_{drive}
\end{equation*}
where $\lambda$ is a factor for weighting the two criteria.
We define the sum of these two items (with $\lambda=1$) to be the inconvenience time for a user.
To achieve this, the reward function is defined as 
\begin{equation*}
    R(t) = KN_{arrive}- N_{wait} - \lambda N_{drive}
\end{equation*}
Where $N_{arrive}, N_{wait}$ and $N_{drive}$ are defined as the number of users who arrived at a station, waited there or were driving at the current timestep, respectively.
$K$ is the reward that the agent receives for successfully completing a dispatch. 
Trivially, as $\gamma \rightarrow 1$, maximizing this reward corresponds to minimizing the previous equation.

\textit{Transition Function: } In this study, the transition function is modeled by a simulation, which is further described in the `Analysis \& Results' section. 
In practice, the transition function will be the real-world movement of cars to and from stations, as well as the updated information that the agent receives from these actions.
Note that this simulation assumes that there are many vehicles that leave and enter the station outside of our control, and these vehicles must be considered part of the transition function.
Once a car recommended by our station begins charging, we have no further influence over it and it becomes part of the environment.


However, he above formulation does not quite fulfill the requirements for a Markov decision process, which requires that the transition function is solely a function of the current state, because of several reasons. 
First, how many cars enter and leave a given charging station may be influenced by how long those cars have already been charging.
In the analysis section, we show that, counter-intuitively, this is not the case.

Second, how many cars enter a station is partially influenced by our previous recommendations. 
To account for this, we add an additional engineered feature included in the station information ($s_n$).
For each $\delta t \in [1, k]$, we include information about the number of cars that are estimated to arrive at timestep $t+\delta t$ as a result of our recommendations.
As this information can be updated by the agent every timestep {\it and} whenever it makes a decision, it allows the agent to remember its queued decisions that it hasn't executed yet.
As a result of this, it is able to coordinate between the many decisions it can be forced to make at each timestep. Koenig et Al. show that an optimal policy learned this way will obtain no less than half of the rewards of an optimal policy that considers all $\mathcal{Q}(t)$ at once, while avoiding the exponential complexity of such an algorithm.

\textit{Classical Model: }A classical method of solving such problems is to use a feed forward Deep Q-Network (DQN). 
In this case, the state is input as a concatenation of all of the parts of the station $s = \{s_n \forall n, s_i \forall i \in \mathcal{Q}(t), s_G\}$. 
The DQN, parameterized by $\theta$, outputs a vector $\vec{q}$, such that $\vec{q}_n$ denotes the Q-value for action $a_n$.
The policy $\pi$, is to then usually select the station with the highest Q-value. 
However, to incentivize exploration, the agent sometimes chooses a random agent with probability $\epsilon$.
The network is then trained via temporal differencing, such that the update is calculated as follows:
\begin{equation*}
    \Delta \theta = \eta  \nabla_\theta\Big(R(t) + \max_{a' \in A}Q(s',a'| \Theta') - Q(s,a| \Theta)\Big)
\end{equation*}
To improve convergence, a replay memory buffer is used to sample old experiences to perform the TD-update on.

\textit{Convolutional Model: }There can be hundreds of stations, meaning that this single network must take in all of the hundreds on inputs at once and map this to the hundreds of appropriate Q-values. 
However, he occupancy has of a station on the southern side of the city likely should have little influence on the agent's evaluation of an arbitrary station on the other side of the city. 
Furthermore, the evaluation process should be roughly the same for each station - look at how far it is, look at whether or not we anticipate a demand in the area, etcetera.
As a result, the network can be improved by evaluating each station individually.
In this case, rather than concatenating all of the states, when calculating $Q(s,a_n)$ for the $i^{th}$ query, the agent is only input with the values of $\{s_n, s_i, s_G\}$.
In this formulation, the evaluation of all of the stations shares the same parameters.
This also enables the agent to operate with far fewer parameters, speeding up computation.
Mathematically, this operation is equivalent to inputting the state as an $Nx1$ image, with depth equal to the number of features for each station plus the number of features for the global state.
Processing each station one at a time then corresponds to doing 1x1 convolution over the stations, hence the name `convolutional model'.

\textit{Graphical Model: }Despite these advantages, this analysis can be slightly reductive. 
Evaluating each station in isolation could potentially miss out on important information - if all other nearby stations are full, the remaining station with vacancies can expect a lot of their demand to be redirected to it.
It is then useful for the station to have a method to view information about nearby states.

Given that the spread of stations is non-uniform, traditional convolutional methods do not work. As a result, this paper employs graph convolution to add context \cite{GCN}.
Specifically, this paper employs a form of graph convolutional networks similar to GraphSage.
Specifically, we define a graph $G = \{V,E,W\}$ over the state space. 
Specifically, we let the set of vertices $V$ to be the set of stations and add edges between two nodes if the corresponding stations are within distance $D$ from each other.
Letting $d(n,m)$ be the distance between stations $n$ and $m$, the corresponding weight is defined as $\alpha_nm e^{-\beta d(n,m)}$, where $\alpha$ is a hyperparameter.

\begin{figure}
    \centering
    \includegraphics[width=0.75\linewidth, angle=270]{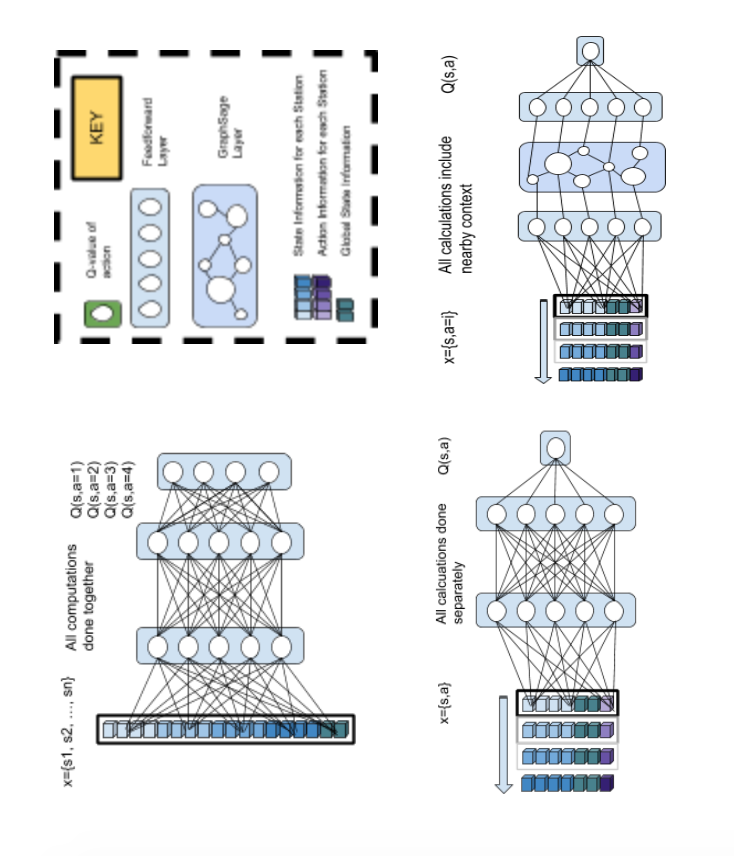}
    \caption{Depictions of easy model type.  \textit{Top-Left:} Classical FFDQN network \textit{Bottom-Left:} Convolutional Model {\it Bottom-Right:} GraphSage Model}
    \label{fig:models}
\end{figure}

Denoting $E(n)$ as the stations that share an edge with $n$, the Q-values of action $n$ are computed as follows.
\begin{equation*}
    h_n = f(s, a_n|\theta_1)
\end{equation*}
\begin{equation*}
    x_n = \{h_n, \sum_{m \in E(n)} \alpha_m h_m\}
\end{equation*}
\begin{equation*}
    q_n = g(x_n | \theta_2)
\end{equation*}
where $f, g$ are feed-forward networks. A diagram explaining each of the three model types is provided in figure (\ref{fig:models}) and the pseudocode provides an overview of the sequential decision process.
\begin{algorithm}[tb]
\caption{Code for agent actions}
\label{code:action}
\textbf{ACT($\pi$, $S$, $\mathcal{Q}(t)$, $E$)}
\begin{algorithmic}[1] 
\STATE SHUFFLE($\mathcal{Q}(t)$)
\FOR{$q^{(j)} \in \mathcal{Q}(t)$}
\FOR{$s^{(i)} \in \text{NEARBY}(q^{(j)}, S)$}
    \STATE bid$^{(i)} = Q(s^{(i)}, q^{(j)}, E)$
\ENDFOR
\STATE to = $\text{argmax}_{j} \Big[\text{bid}^{(i)}\Big]$
\STATE dist = DISTANCE($s^{(to)}$, $q^{(j)}$)
\STATE $s^{(i)}_{d_{dist}} += 1$
\ENDFOR
\end{algorithmic}
\label{choose}
\end{algorithm}

An alternate way to formulate the problem is as a multi-agent problem.
In this formulation, each station could be formulated as its own agent, and for each query, it has a decision to accept or not accepted the user to be recommended to it.
In this way, each station receives reward for the queries that are dispatched to it.
However, our MDP requires that the agent can only recommend one station at a time, so there needs to be consensus between stations, such that exactly one station `accepts' each query.

Let $Q_n(s, 1)$ be the discounted expected reward of station $n$ if it accepts, and $Q_n(s,0)$ if it rejects.
Then maximizing the total reward over all stations is maximized by recommending the user to station $n$ such that \begin{equation*}
    n = \text{argmax}_{n \in N} Q_n(s,1) - Q_n(s,0)
\end{equation*}
We explore this model in analysis and results below.

To make any of these models learn anything at all, several small improvements are also necessary.
First, the agent can only recommend agents to stations within distance $D$. 
If there are less than $5$ stations within distance $D$, the agent can recommend to the 5 nearest stations instead.
Additionally, to stabilize training, this paper uses the Double DQN architecture, as pioneered in \cite{DDQN} .
Finally, a dueling DQN architecture was implemented, using a convolutional neural network to learn the value function.
\section{Experiment Design}
\begin{figure}
    \centering
    \includegraphics[width=0.75\linewidth]{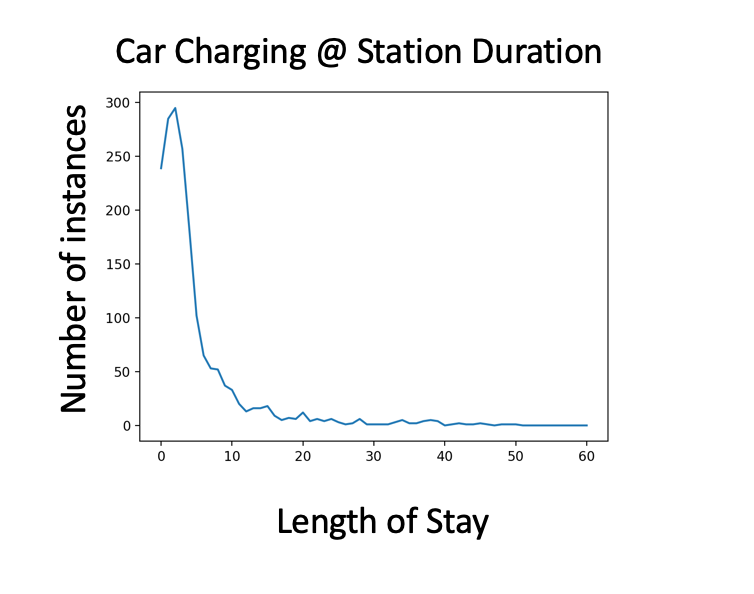}
    \caption{Counts of charging durations by number of observations}
    \label{duration}
\end{figure}

{\it Data Analysis: }The algorithm is trained and tested in a simulation based on data from over 6 months of data. 
As noted above, the region is divided in to a rectangular grid. In the case of Beijing, the city is divided in to a grid of size $144 \times 126$.
Time is discretized in to non-overlapping periods, each with duration $\tau=15$ minutes.
Charger data is collected 514 chargers across 73 stations and 96,000 timesteps.
This is used to measure how long cars stay at a station by measuring the time between when when a charger becomes occupied and when it first becomes free again. 
From this, a probability distribution of charging durations is derived, as shown in figure (\ref{duration}).
We note that the distribution appears to roughly follow a geometric distribution (p < 5e-4), which has the convenient property of being memoryless. 
This means that knowing how long a vehicle has been charging for does not help predict how much longer it will continue charging, so the model does not need to keep track of this information.

\begin{figure}
    \centering
    \includegraphics[width=0.75\linewidth]{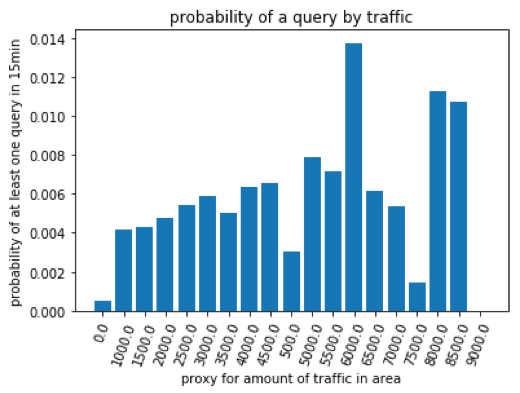}
    \caption{The number of queries received in an area doesn't correlate strongly with the number of cars in the area. Note that there is higher variance for higher-traffic data values, as there are fewer datapoints.}
    \label{traffic}
\end{figure}

As a result, at each timestep, we can model the number of remaining vehicles as a geometric distribution.
The next question is the number of incoming cars.
Naturally, one would expect that the number of cars coming in to the station and the number of cars querying would be proportional to the traffic in the surrounding area. 
However, this assumption does not match the data, as shown in figure (\ref{traffic}). 
As nearby traffic doesn't seem to correlate strongly with any core pieces of the environment, it is also not included in the state.

{\it Simulation: }With these considerations, the simulation is defined as follows. 
The agent is trained in an episodic manner, with each simulation starting and ending at midnight. 
The number of cars that remain at the station at each timestep is sampled from a geometric distribution. 
From there, a number of cars incoming from outside for the system is sampled from a distribution modeled off of historical data.
Finally, the dispatches made by our system are handled.
If possible, all of the vehicles that are within 1 timestep of their target station move to that station and begin charging. 
If there are not enough chargers, then those vehicles that couldn't be accommodated wait until the next timestep (at which point they will get priority over arriving vehicles).
Finally, the position of all other vehicles is updated. 
A summary of this algorithm is provided in pseudo code.


\begin{algorithm}[tb]
\caption{Code for training simulation}
\label{code:simulation}
\textbf{SIMULATE($f$, $g$, $\pi$, $S$)}
\begin{algorithmic}[1] 
\STATE Let $t=0$.
\WHILE{$t < T_{END}$}
\FOR{$s^{(i)} \in S$}
\STATE $N_{cars}^{(i)}(t+1) = N^{(i)}_{rem} + N^{(i)}_{inc} + N^{(i)}_{rec}$
\IF{$s^{(i)}_T < N_{cars}^{(i)}(t+1)$}
    \STATE $s^{(i)}_{d_0} += N_{cars}^{(i)}(t+1) - s^{(i)}_T$
    \STATE $N_{cars}^{(i)}(t+1) = s^{(i)}_T$
\ENDIF
\STATE $s^{(i)}_O = s^{(i)}_T - N_{cars}^{(i)}(t+1)$
\ENDFOR
\STATE $\mathcal{Q}(t) \sim g(t, dow)$
\STATE ACT($\pi$, $S$, $\mathcal{Q}(t)$)
\ENDWHILE
\end{algorithmic}
\end{algorithm}
\section{Results \& Analysis}
We evaluate the proposed models in the simulation defined above. The convolutional and graphical models are build with a small network size of only 100 nodes per hidden layer and 3 hidden layers.
A variety of network sizes were tried for the classical DQN, but the best performing network was substantially larger, with 250 nodes per hidden layer. The exploration probability, $\epsilon$, is initially set at 0.9 and decreases exponentially to 0.1 throughout training.

\begin{figure}
    \centering
    \includegraphics[width=0.75\linewidth]{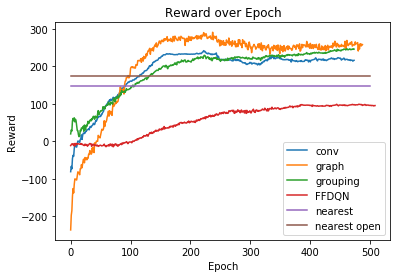}
    \caption{Convolutional and Graphical Models Substantially Outperform Other Strategies}
    \label{fig:rewards}
\end{figure}

Aside from the classic DQN model, we evaluate our model against several rule-based metrics, defined as follows:
\begin{itemize}
    \item \textbf{Nearby:} Always recommends the user to the nearest station in a greedy manner. This should minimize $t_{drive}$ for all users. 
    \item \textbf{Open:} Always recommends the user to the station that has the highest number of open chargers, breaking ties by distance. This would ideally minimize $t_{wait}$ for all users.
    \item \textbf{NearestOpen:} An approximation of the method described in \cite{RuleBasedInvalid}, but avoids using future information. Will usually choose nearest station, but will choose other stations if the desired station is full.
\end{itemize}
We test 4 different deep-learning based methods: Graphical, Convolutional and Classical DQN are all as described above. `Grouping' denotes a convolutional model that first recommends the user to a group of stations and then uses greedy nearby recommendation once the user arrives. Figure (\ref{fig:rewards}) shows a plot of the average rewards received on the test seed by epoch. Note that 'Open' is not shown on graphs because it performs so poorly that it makes the other models hard to compare.

The results shown strongly suggest that classical DQN methods are not able to solve this problem to any satisfying degree, as it performs significantly worse than both rule-based baselines. However, both the convolutional and graphical models substantially outperform the baselines. However, there are a couple of interesting things to note about their performance.

Firstly, there is an interesting phenomenon where the average reward nearly monotonically increases each epoch, but then there is a slight downward dip after roughly epoch 200, before the rewards start climbing upwards again. There are two possible reasons for this. The more interesting reason, is that it may be an instance of a broader phenomenon, called `Deep Double Descent', where neural networks often undergo a period where their performance decreases after an initial rapid decrease \cite{DeepDoubleDescent}. 

The second conceivable reason is a bit more banal - it has to do with random seed. The data is tested on random seeds zero through ten. The agent initially plays the simulation with random seed zero, and each time it plays, it increases the random seed by one. Because of this, it does initially see the random seeds that it is eventually tested on, once for each. However, this experiment uses a experience replay buffer to stabilize convergence, and epoch 200 is roughly where those `memories' would leave the buffer. It is fairly dubious that this effect, if present, is very strong, because the policy that the agent followed on its first attempt is likely very different from its final policy, but it is something to be aware of for future experiments.

The other thing to note is that the graphical model does indeed outperform the baseline convolutional model, but it's not the only one to do so. Adding grouping to the convolutional model appears to achieve roughly the same final results, although it doesn't converge as quickly. Future work will involve testing whether or not combining the graphical model and the grouping mechanism results in an even better model. The natural initial response to this question might be a hesitant yes, but, when looking at the wait times, it's not totally clear that it is possible to improve.

\begin{figure}
    \centering
    \includegraphics[width=0.75\linewidth]{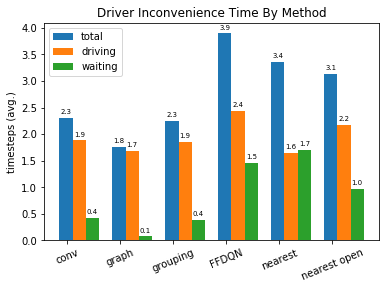}
    \caption{Graphical Model Surprisingly Performs Near-Optimally}
    \label{fig:inconveniences}
\end{figure}

For this, we reference figure (\ref{fig:inconveniences}). We again note that the theoretical minimum drive distance is equal to the drive distance given by the `nearest' recommender. Looking at the graph, we see that the graphical and grouping models are actually very close in this regard - the average drive time is only 0.1 and 0.3 timesteps more than that of the `nearest' recommender. Despite this, the graphical model achieves near-zero wait time. Impressively, the graphing model achieves a result that is nearly as good as a fantasy world where every user drives to the nearest station and there's always a spot for them - it's average wait time is only 0.2 more than that! 

\begin{table*}[ht]
\centering
    \begin{tabular}{c|cccccc}
     & \textbf{Graphical} &  \textbf{Convolutional} & \textbf{FFDQN} & \textbf{Nearest} & \textbf{Open} & \textbf{Nearest Open}\\\hline
    Reward &  \textbf{291} & 242 & 97 & 148 & -910 & 173\\
    Inconvenience Time (m) & \textbf{8.80} & 11.55 & 19.46 & 16.51 & 80.15 & 15.62\\
    Wait Time (m) & \textbf{0.4} & 2.15 & 7.30 & 8.50 & 16.90 & 4.81  \\
    Drive Time(m)  & 8.40 & 9.41 & 12.15 & \textbf{8.00} & 63.25 & 10.80\\
    \end{tabular}
    \caption{Averages of useful metrics in simulation organized by algorithm used for routing. Inconvenience time represents the sum of time spent waiting and driving.}
    \label{fig:resultstable}
\end{table*}

This indicates that, while the graphical model might not be perfect, there likely isn't much to gain from improving it further, as you could only reduce wait times by another 10\% at most. To contrast this, the graphical model reduces wait times by about 45\% over all of the baselines, which is much more substantial. These results are condensed in table (\ref{fig:resultstable}).

However, there are a number of caveats to these statements. The first, and biggest one, is that these are only single runs for each model. Reinforcement learning models are notorious for being unstable, and that has been my experience with this project as well. Its very possible that the random seeds gave good initial weights that helped things converge smoothly. With more resources and time, further runs would prove helpful to demonstrate the effectiveness of this method.

The second caveat is that these results are within a simulation, and it may be significantly easier to optimize the simulation than the real world. This simulation-gap is a common difficulty that many reinforcement learning projects face. One potentially impactful gap is that the nature of the data led the simulation to use a larger timestep than it would use after deployment. Additionally, the simulation did not involve other factors such as inclement weather or unexpected traffic issues.

\begin{figure}
    \centering
    \includegraphics[width=0.75\linewidth]{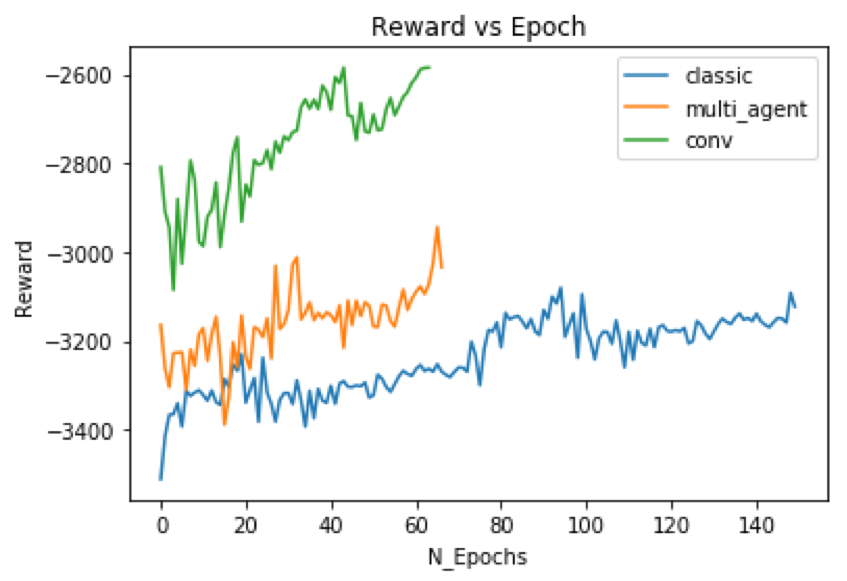}
    \caption{Multi-Agent Model Mysteriously Underperforms}
    \label{fig:ma}
\end{figure}

The last set of experiments to touch upon are the multi-agent experiments. Unlike the graphical model, these heavily underperformed. While it did outperform the baseline, it still wasn't anywhere close to the convolutional model and it trained extremely slowly. It is difficult to solid justification for this, and is worth investigating in the future. A plot of the reward functions is provide in figure (\ref{fig:ma}).

One possible explanation for this behaviour is that the multi-agent model doesn't quite fully satisfy the MDP property. Whenever it does an auction, the model doesn't `know' if it will have another query this timestep or not, even though that is pre-determined, therefore making it suboptimal. That being said, this same effect should apply to all of the other models as well, but those performed fine. Another possibility is that it simply takes longer to converge, or that the non-stationarity of the other agents makes the environment itself non-stationarity. According to \cite{MultiAgent}, the fact that all agents share parameters should prevent that from happening, but it's possible that something still went wrong.
\section{Conclusion and Future Work}
In total, this report presents a framework for significantly improving charging station recommendations in response to dynamic, user-generated queries. The graphical model proposed is the first of its kind, and it balances between providing the canvas for complex interactions to be discovered and limiting the search space to allow it to tackle more difficult problems.

There are two primary directions for future work that immediately stand out. The first is strictly application focused. This model needs to be stress-tested on multiple cities over multiple runs, and could be implemented in a real-world system. To do so, we would likely want to collect more statistics, such as what percentage of users have to wait, and how long the longest wait times are. There are always more experiments to be run.

The other direction is the most theoretical, but has a lot of potential. In theory, everything done by this network could be down by a fully-connected network, but no network ever came close to achieving these results. As I see it, feature-engineering and simulation aside, all that this paper did is tie weights together and prune weights in the fully-connected network. It would be wonderful if there were a system that could detect the pattern in the inputs (that every $k$ dimensions represented a station) and automatically do this pruning and weight-tying. This direction still involves a large amount of work before coming to fruition, but it has many possible applications.

\bibliographystyle{named}
\bibliography{bibliography.bib}

\begin{thebibliography}{}

\bibitem[\protect\citeauthoryear{Alabbasi \bgroup \em et al.\egroup
  }{2019}]{Taxi5}
Abubakr Alabbasi, Arnob Ghosh, and Vaneet Aggarwal.
\newblock Deeppool: Distributed model-free algorithm for ride-sharing using
  deep reinforcement learning.
\newblock {\em arXiv preprint arXiv:1903.03882}, 2019.

\bibitem[\protect\citeauthoryear{Defferrard \bgroup \em et al.\egroup
  }{2016}]{ChebNets}
Micha{\"e}l Defferrard, Xavier Bresson, and Pierre Vandergheynst.
\newblock Convolutional neural networks on graphs with fast localized spectral
  filtering.
\newblock In {\em Advances in neural information processing systems}, pages
  3844--3852, 2016.

\bibitem[\protect\citeauthoryear{Foerster \bgroup \em et al.\egroup
  }{2017}]{MultiAgent}
Jakob Foerster, Nantas Nardelli, Gregory Farquhar, Triantafyllos Afouras,
  Philip~HS Torr, Pushmeet Kohli, and Shimon Whiteson.
\newblock Stabilising experience replay for deep multi-agent reinforcement
  learning.
\newblock In {\em Proceedings of the 34th International Conference on Machine
  Learning-Volume 70}, pages 1146--1155. JMLR. org, 2017.

\bibitem[\protect\citeauthoryear{He and Shin}{2019}]{GoodImages}
Suining He and Kang~G Shin.
\newblock Spatio-temporal capsule-based reinforcement learning for
  mobility-on-demand network coordination.
\newblock In {\em The World Wide Web Conference}, pages 2806--2813. ACM, 2019.

\bibitem[\protect\citeauthoryear{Houbbadi \bgroup \em et al.\egroup
  }{2019}]{EBus}
Adnane Houbbadi, Rochdi Trigui, Serge Pelissier, Eduardo Redondo-Iglesias, and
  Tanguy Bouton.
\newblock Optimal scheduling to manage an electric bus fleet overnight
  charging.
\newblock {\em Energies}, 12(14):2727, 2019.

\bibitem[\protect\citeauthoryear{Jindal \bgroup \em et al.\egroup
  }{2018}]{Taxi1}
Ishan Jindal, Zhiwei~Tony Qin, Xuewen Chen, Matthew Nokleby, and Jieping Ye.
\newblock Optimizing taxi carpool policies via reinforcement learning and
  spatio-temporal mining.
\newblock In {\em 2018 IEEE International Conference on Big Data (Big Data)},
  pages 1417--1426. IEEE, 2018.

\bibitem[\protect\citeauthoryear{Kipf and Welling}{2016}]{GCN}
Thomas~N Kipf and Max Welling.
\newblock Semi-supervised classification with graph convolutional networks.
\newblock {\em arXiv preprint arXiv:1609.02907}, 2016.

\bibitem[\protect\citeauthoryear{Mei and Montanari}{2019}]{DeepDoubleDescent}
Song Mei and Andrea Montanari.
\newblock The generalization error of random features regression: Precise
  asymptotics and double descent curve.
\newblock {\em arXiv preprint arXiv:1908.05355}, 2019.

\bibitem[\protect\citeauthoryear{Mnih \bgroup \em et al.\egroup }{2013}]{Mnih1}
Volodymyr Mnih, Koray Kavukcuoglu, David Silver, Alex Graves, Ioannis
  Antonoglou, Daan Wierstra, and Martin Riedmiller.
\newblock Playing atari with deep reinforcement learning.
\newblock {\em arXiv preprint arXiv:1312.5602}, 2013.

\bibitem[\protect\citeauthoryear{Mnih \bgroup \em et al.\egroup
  }{2015}]{MnihHuman}
Volodymyr Mnih, Koray Kavukcuoglu, David Silver, Andrei~A Rusu, Joel Veness,
  Marc~G Bellemare, Alex Graves, Martin Riedmiller, Andreas~K Fidjeland, Georg
  Ostrovski, et~al.
\newblock Human-level control through deep reinforcement learning.
\newblock {\em Nature}, 518(7540):529, 2015.

\bibitem[\protect\citeauthoryear{Oda and Joe-Wong}{2018}]{Taxi2}
Takuma Oda and Carlee Joe-Wong.
\newblock Movi: A model-free approach to dynamic fleet management.
\newblock In {\em IEEE INFOCOM 2018-IEEE Conference on Computer
  Communications}, pages 2708--2716. IEEE, 2018.

\bibitem[\protect\citeauthoryear{Oda and Tachibana}{2018}]{Taxi3}
Takuma Oda and Yulia Tachibana.
\newblock Distributed fleet control with maximum entropy deep reinforcement
  learning.
\newblock 2018.

\bibitem[\protect\citeauthoryear{Pineda}{1987}]{GNNRecur}
Fernando~J Pineda.
\newblock Generalization of back-propagation to recurrent neural networks.
\newblock {\em Physical review letters}, 59(19):2229, 1987.

\bibitem[\protect\citeauthoryear{Ramachandran \bgroup \em et al.\egroup
  }{2018}]{StationPlanning}
Anshul Ramachandran, Ashwin Balakrishna, Peter Kundzicz, and Anirudh Neti.
\newblock Predicting electric vehicle charging station usage: Using machine
  learning to estimate individual station statistics from physical
  configurations of charging station networks.
\newblock {\em arXiv preprint arXiv:1804.00714}, 2018.

\bibitem[\protect\citeauthoryear{Russell and Norvig}{2016}]{RusselNorvig}
Stuart~J Russell and Peter Norvig.
\newblock {\em Artificial intelligence: a modern approach}.
\newblock Malaysia; Pearson Education Limited,, 2016.

\bibitem[\protect\citeauthoryear{Scarselli \bgroup \em et al.\egroup
  }{2008}]{OGGNN}
Franco Scarselli, Marco Gori, Ah~Chung Tsoi, Markus Hagenbuchner, and Gabriele
  Monfardini.
\newblock The graph neural network model.
\newblock {\em IEEE Transactions on Neural Networks}, 20(1):61--80, 2008.

\bibitem[\protect\citeauthoryear{Shou \bgroup \em et al.\egroup }{2019}]{Taxi4}
Zhenyu Shou, Xuan Di, Jieping Ye, Hongtu Zhu, and Robert Hampshire.
\newblock Where to find next passengers on e-hailing platforms?-a markov
  decision process approach.
\newblock {\em arXiv preprint arXiv:1905.09906}, 2019.

\bibitem[\protect\citeauthoryear{Sutton and Barto}{2018}]{SuttonBarto}
Richard~S Sutton and Andrew~G Barto.
\newblock {\em Reinforcement learning: An introduction}.
\newblock MIT press, 2018.

\bibitem[\protect\citeauthoryear{Tian \bgroup \em et al.\egroup
  }{2016}]{RuleBasedInvalid}
Zhiyong Tian, Taeho Jung, Yi~Wang, Fan Zhang, Lai Tu, Chengzhong Xu, Chen Tian,
  and Xiang-Yang Li.
\newblock Real-time charging station recommendation system for electric-vehicle
  taxis.
\newblock {\em IEEE Transactions on Intelligent Transportation Systems},
  17(11):3098--3109, 2016.

\bibitem[\protect\citeauthoryear{Valogianni \bgroup \em et al.\egroup
  }{2013}]{Charging1}
Konstantina Valogianni, Wolfgang Ketter, and John Collins.
\newblock Smart charging of electric vehicles using reinforcement learning.
\newblock In {\em Workshops at the Twenty-Seventh AAAI Conference on Artificial
  Intelligence}, 2013.

\bibitem[\protect\citeauthoryear{Verma \bgroup \em et al.\egroup
  }{2017}]{RegionSplit}
Tanvi Verma, Pradeep Varakantham, Sarit Kraus, and Hoong~Chuin Lau.
\newblock Augmenting decisions of taxi drivers through reinforcement learning
  for improving revenues.
\newblock In {\em Twenty-Seventh International Conference on Automated Planning
  and Scheduling}, 2017.

\bibitem[\protect\citeauthoryear{Xiong \bgroup \em et al.\egroup
  }{2018}]{Charging2}
Rui Xiong, Jiayi Cao, and Quanqing Yu.
\newblock Reinforcement learning-based real-time power management for hybrid
  energy storage system in the plug-in hybrid electric vehicle.
\newblock {\em Applied energy}, 211:538--548, 2018.

\bibitem[\protect\citeauthoryear{Xu \bgroup \em et al.\egroup }{2018}]{DidiBig}
Zhe Xu, Zhixin Li, Qingwen Guan, Dingshui Zhang, Qiang Li, Junxiao Nan,
  Chunyang Liu, Wei Bian, and Jieping Ye.
\newblock Large-scale order dispatch in on-demand ride-hailing platforms: A
  learning and planning approach.
\newblock In {\em Proceedings of the 24th ACM SIGKDD International Conference
  on Knowledge Discovery \& Data Mining}, pages 905--913. ACM, 2018.

\bibitem[\protect\citeauthoryear{Zhao \bgroup \em et al.\egroup
  }{2018}]{Charging4}
Pu~Zhao, Yanzhi Wang, Naehyuck Chang, Qi~Zhu, and Xue Lin.
\newblock A deep reinforcement learning framework for optimizing fuel economy
  of hybrid electric vehicles.
\newblock In {\em 2018 23rd Asia and South Pacific Design Automation Conference
  (ASP-DAC)}, pages 196--202. IEEE, 2018.

\bibitem[\protect\citeauthoryear{Zhou \bgroup \em et al.\egroup
  }{2019}]{SpatialDistribution}
Ming Zhou, Jiarui Jin, Weinan Zhang, Zhiwei Qin, Yan Jiao, Chenxi Wang, Guobin
  Wu, Yong Yu, and Jieping Ye.
\newblock Multi-agent reinforcement learning for order-dispatching via
  order-vehicle distribution matching.
\newblock In {\em Proceedings of the 28th ACM International Conference on
  Information and Knowledge Management}, pages 2645--2653. ACM, 2019.

\bibitem[\protect\citeauthoryear{Zou \bgroup \em et al.\egroup
  }{2016}]{Charging3}
Yuan Zou, Teng Liu, Dexing Liu, and Fengchun Sun.
\newblock Reinforcement learning-based real-time energy management for a hybrid
  tracked vehicle.
\newblock {\em Applied energy}, 171:372--382, 2016.

\end{thebibliography}

\section{Appendix}
Calculations on time savings done using the assumption that the average individual drives 13,000 miles per year and that the average electric vehicle range is 110 miles (comparable to the Nissan Leaf or BAIC EC180). Of 560,000 electric vehicles on the road worldwide, if half of them see use, that results in 33 million fill-ups per year. Our simulation estimates a 7.7 minute reduction in average wait time per fill-up.

\end{document}